\documentclass[lettersize,journal]{IEEEtran}
\usepackage{amsmath,amsfonts}
\usepackage{algorithmic}
\usepackage{algorithm}
\usepackage{array}
\usepackage[caption=false,font=normalsize,labelfont=sf,textfont=sf]{subfig}
\usepackage{textcomp}
\usepackage{stfloats}
\usepackage{url}
\usepackage{verbatim}
\usepackage{graphicx}
\usepackage{cite}
\usepackage{xcolor}
\usepackage{hyperref}
\usepackage{soul}
\newcommand{\jpr}[1]{\textcolor{blue}{#1}}
\newcommand{\comments}[1]{}

\begin{document}

\title{StereoTac: a Novel Visuotactile Sensor that Combines Tactile Sensing with 3D Vision}

\author{Etienne Roberge$^{1}$, Guillaume Fornes$^{2}$, Jean-Philippe Roberge$^{1}$
\thanks{$^{1}$Etienne Roberge and Jean-Philippe Roberge are with the Command and Robotics Laboratory, École de technologie supérieure, Montreal, Quebec, H3C1K3, Canada (e-mail: etienne.roberge.1@ens.etsmtl.ca; jean-philippe.roberge@etsmtl.ca).}
\thanks{$^{2}$Guillaume Fornes is with the Bordeaux Institute of Technology, ENSEIRB-MATMECA, 1 avenue du Dr Albert Schweitzer B.P. 99 33402 Talence,
France (e-mail: gfornes@enseirb-matmeca.fr).}
\thanks{Supported by the Natural Sciences and Engineering Research Council of Canada (NSERC) under grant award RGPIN-2022-04884.}
}


\markboth{}
{Roberge \MakeLowercase{\textit{et al.}}: StereoTac} 

\maketitle

\begin{abstract}
    Combining 3D vision with tactile sensing could unlock a greater level of dexterity for robots and improve several manipulation tasks. However, obtaining a close-up 3D view of the location where manipulation contacts occur can be challenging, particularly in confined spaces, cluttered environments, or without installing more sensors on the end effector.\comments{However, it is often difficult to have a close-up 3D view of the location where manipulation contacts will occur, especially in confined spaces and cluttered environments, or without having to install additional sensors on the end effector.} In this context, this paper presents \textit{StereoTac}, a novel vision-based sensor that combines tactile sensing with 3D vision. The proposed sensor relies on stereoscopic vision to capture a 3D representation of the environment before contact and uses photometric stereo to reconstruct the tactile imprint generated by an object during contact. To this end, two cameras were integrated in a single sensor, whose interface is made of a transparent elastomer coated with a thin layer of paint with a level of transparency that can be adjusted by varying the sensor's internal lighting conditions. We describe the sensor's fabrication and evaluate its performance for both tactile perception and 3D vision. Our results show that the proposed sensor can reconstruct a 3D view of a scene just before grasping and perceive the tactile imprint after grasping, allowing for monitoring of the contact during manipulation.\comments{We show that the proposed sensor can be used to reconstruct a 3D view of a scene just before grasping, and perceive the tactile imprint after grasping, which allows monitoring of the contact during manipulation.}
\end{abstract}


\section{Introduction}
\IEEEPARstart{T}{actile} sensing is often an important capability for robots that interact with their environment and that perform tasks such as grasping, manipulating and assembling objects. In recent years, significant progress has been made in developing tactile sensors that can assist robots with determining a myriad of physical attributes related to the objects to manipulate such as their pose, shape and even their texture at a very fine scale. These sensors, which are made from various technologies including piezoelectric, capacitive, and optical sensing, can help robots to handle a wide range of objects\textemdash from small and delicate to large and irregular\textemdash which, by extension, contribute to increase their overall dexterity. When complementing vision, tactile sensing can also unlock the ability of achieving new, complex tasks that require additional information not provided by either sensing modality alone.

\begin{figure}[thpb]
      \centering
      \includegraphics[angle=0, width=0.84\linewidth]{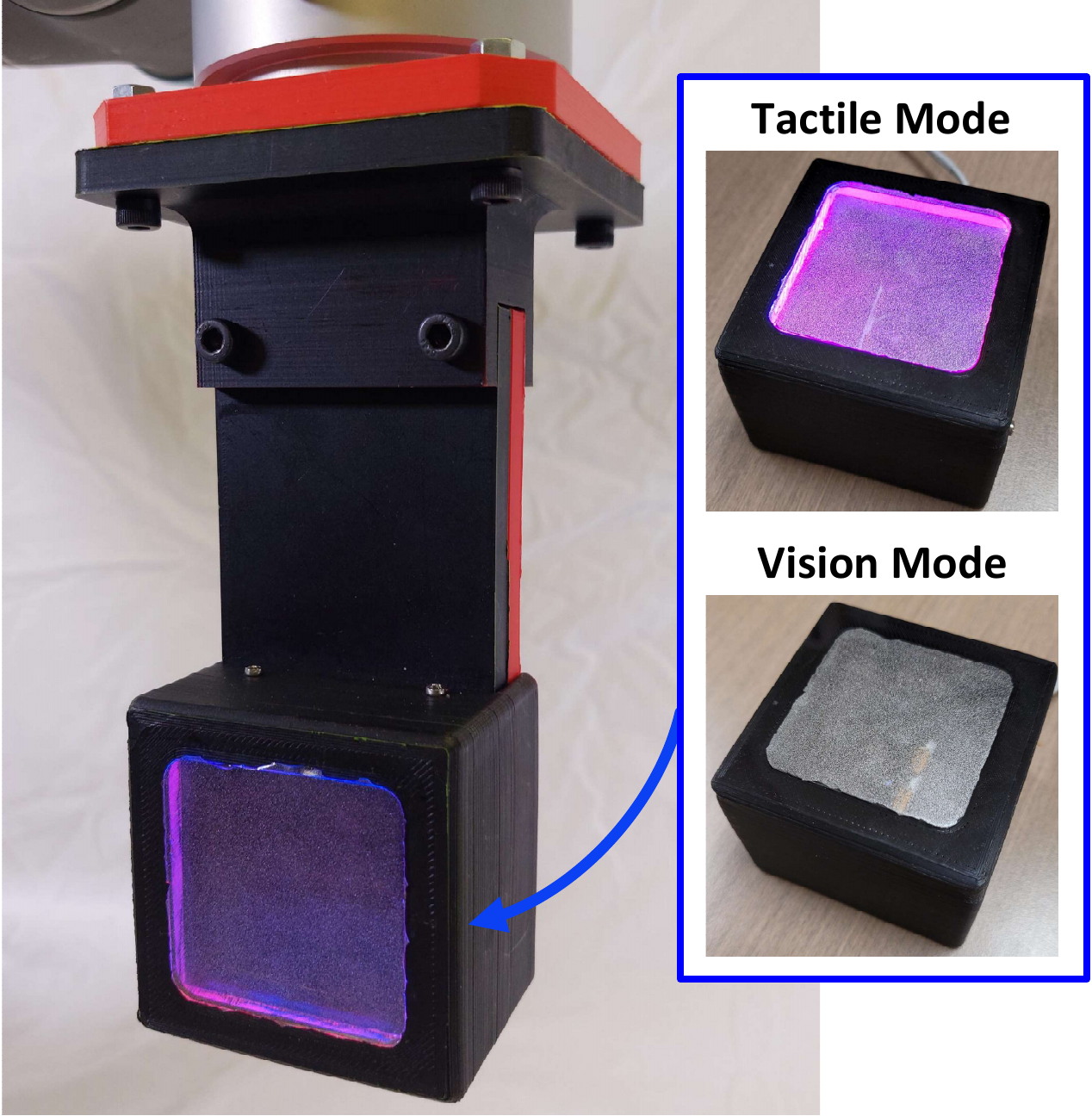}\vspace{-0.12in}
      \caption{A photo showing \textit{StereoTac}, a visuotactile sensor combining photometric stereo with stereoscopic vision. }\vspace{-0.12in}
      \label{fig:sensorPhoto}
      \vspace{-10pt}
\end{figure}

However, a common challenge in tactile sensing and visual perception during robotic manipulation is the reduced object visibility that will typically occur when a robot reaches for an object and/or when the gripper encompasses it as it performs its grasp. This is particularly true when the manipulator is used for reaching objects in confined spaces, such as cabinets or boxes, or when manipulation occurs in cluttered environments. In these scenarios, the robot will often create obstructions to cameras that are statically affixed to the robotic cell. This can limit the accuracy and reliability of the grasp, as the vision system cannot continuously and unobstructedly track the object, which may however be prone to moving as the robot approaches. Alternatively, vision system(s) can be mounted directly to the robot wrist to get a closer perspective to where manipulation contacts will be generated. However, this generally increases the size of the tool at the end effector, which can in turn lead to additional constraints on movement / reduced motion and an overall augmented bulkiness of the system. This is particularly true when 3D vision is needed\textemdash even though small 3D vision systems exist, time-of-flight, structured light and stereoscopic vision to name a few, will generally require more internal components and more space than what is typically needed by 2D cameras. Combining 3D vision with tactile sensing could be an advantage in several manipulation tasks, but acquiring complete, unobstructed tridimensional view of the object close to where manipulation actually happens is often still considered a challenge.

\begin{figure*}[t!]
    \centering
    \includegraphics[width=\textwidth]{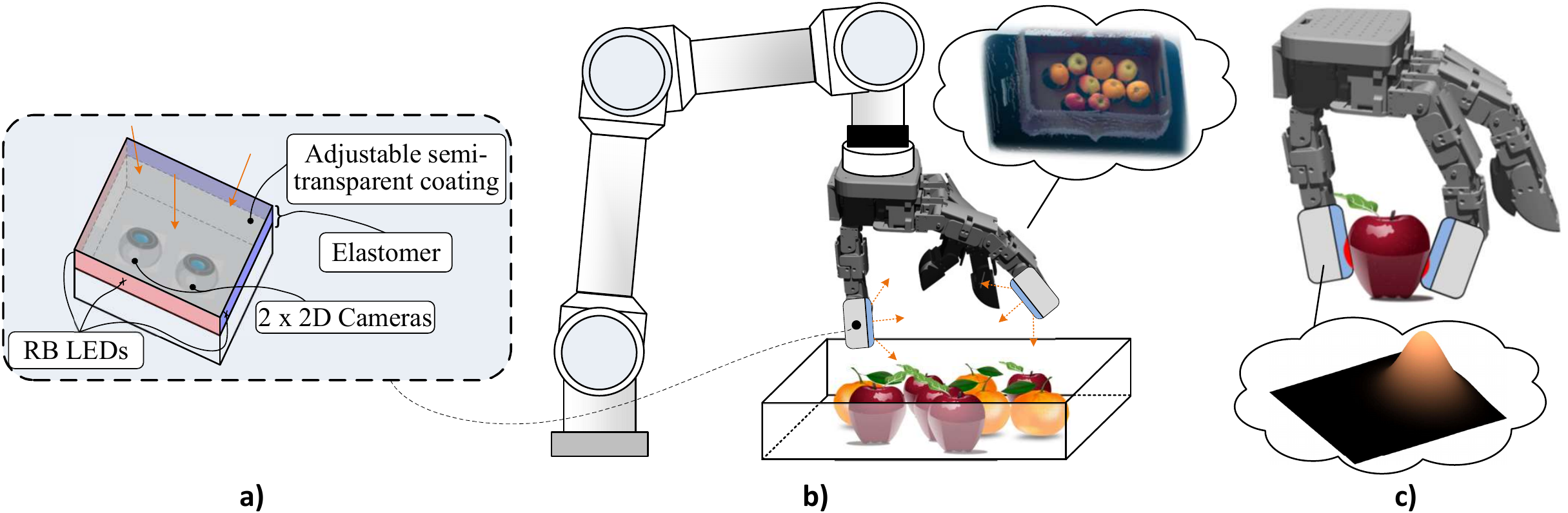}\vspace{-0.15in}
    \caption{An example of a use case scenario with StereoTac. a) StereoTac's main components; b) Here, two sensors are integrated to robotic hand and provide a stitched 3D close-up view for grasp planning; c) After an objet is picked up, the sensor allows the reconstruction of the tactile imprint.}\vspace{-0.2in}
    \label{fig:SensorExampleScenario}
\end{figure*}

To address this issue, we propose a novel vision-based tactile sensor that combines stereoscopic vision (3D vision) with photometric stereo (tactile), as displayed in Fig.~\ref{fig:sensorPhoto}. The proposed sensor uses a semi-transparent soft skin and two cameras to enable high-resolution, multi-modal tactile sensing for robotic manipulation. The sensor is inspired by the idea of \textit{whole-body vision}~\cite{Yamaguchi2017Optical} which consists of having the ability to perceive tactile stimuli as well as being able to see through the skin. In our proposed implementation, the contact interface is made of a transparent elastomer covered with a thin layer of reflective paint, which allows two cameras to capture the deformation of the surface induced by the contact while still being able to see through the skin by varying internal lighting conditions. This allows the sensor to acquire 3D images of the scene, even when the manipulator is close (from 5 to 60 cm) from the grasping site, and to get tactile imprints after contacting the object, as illustrated in Fig.~\ref{fig:SensorExampleScenario}.
\comments{
Using a monocular camera view as a new modality for tactile sensor provides valuable information, but it suffers from several limitations that can affect the accuracy and reliability of the grasping process. One major limitation is the lack of depth information, which makes it difficult for the robot to estimate the proximity, 6D pose and size of the object. This can lead to grasping failures, especially for objects with thin or irregular shapes. Stereoscopic vision, on the other hand, can provide accurate depth information by comparing the difference in the object's position from two cameras with slightly different viewpoints. This could significantly improve the accuracy and reliability of the grasping process, especially in unstructured and cluttered environments.

\jpr{ We propose a novel tactile sensor built upon Finger-STS \cite{Hogan2022Finger}}, which uses a transparent elastomer coated with a reflective layer to reflect light and capture the deformation of the membrane. By analyzing the patterns of light and shadow on the deformed membrane, the sensor can measure the 3D shape and surface normal of the contact surface, and relate it to the 3D pose and size of the contacted object. Furthermore, the addition of a second camera allows for the creation of stereo capabilities, enabling the system to perform more accurately and be adaptive in dynamic and cluttered environments.
}
The main contributions of this paper are:
\begin{itemize}
\item The design of \textit{StereoTac}, a novel visuotactile sensor that combines tactile sensing with 3D vision. To the best of our knowledge, it is the first time a vision-based tactile sensor that uses photometric stereo with stereoscopic vision was designed to capture both visual (3D) and tactile data from the same location;
\comments{\item  A general approach to the 3D reconstruction of the tactile imprint generated by an object in contact with a semi-transparent elastomer\comments{ as well as insights about important parameters to tune for acquiring 3D representations of the environment before making contact}; }
\item An analysis of the impact the level of transparency has on both tactile imprint reconstruction and 3D perception;
\item Experimental results that demonstrate the sensor's ability to capture both tactile and 3D vision. The stereoscopic camera is compared to a similar off-the-shelf product and the whole sensor is integrated to a robotic arm for 3D object reconstruction (shown in the \href{https://youtu.be/64nJgx2AMV4}{\textcolor{blue}{{\setul{1pt}{.4pt}\ul{accompanying video}}}}).
\end{itemize}

\comments{\item The \textbf{evaluation} of the effect of \jpr{STS technology} on both tactile and stereoscopic data. \jpr{We evaluate the performance of the STS sensor} in comparison to the current generation of short-range stereoscopic sensor available on the market and compare the use of \jpr{STS membrane} and traditional opaque elastomer for tactile sensing. The experiments introduced demonstrate the advantages of using \jpr{STS technology} in visuotactile sensing and provide insight into the trade-offs between visual and tactile sensing.}

In the next sections, we first provide background information about vision-based sensors, with a focus on sensors that have the ability to acquire external vision as well as tactile stimuli. In section~\ref{sec:DesignFabrication}, we discuss StereoTac's operating principles as well as its fabrication process. In sections~\ref{sec:Stereo} and~\ref{sec:tactileSensing}, we present experimental results related to the performance of both tactile and visual data, and use of the sensor in the real world. Finally, we discuss the main results this work generated as well as future work opportunities. 

\section{Related Work}\label{sec:RelatedWork}\vspace{-0.05in}
\subsection{Vision-Based Sensors}\vspace{-0.05in}
\comments{Tactile sensors can provide crucial information during manipulation tasks and are often considered as a complementary sensor to vision. Tactile sensors have been shown to be useful for a myriad of tasks such as assessing the quality of a grasp~\cite{Cockbum2017Grasp} or for estimating the pose of an object in the gripper~\cite{Shaoxiong2021GelSight}. Thus, since many decades, }Considerable efforts have been devoted, during the last decades, to the development of high-performance tactile sensors based on different transduction principles. Chi et al.~\cite{Chi2018Recent} review the most common transduction types in tactile sensors, which include piezoresistive, piezoelectric, capacitive, inductive, magnetic, barometric and optical sensors. A subcategory of optical sensors includes those based on vision~\cite{Shimonomura2019Tactile}, sometimes also referred to as "visuotactile sensors". Vision-based tactile sensors have brought significant interest in the last years,\comments{ which is} partly due to the fact that their data processing can leverage well-known computer vision techniques. Furthermore, these sensors benefit from recent progress in the field of cameras and are generally considered as high resolution, affordable and reliable sensors, when broadly compared to other transduction techniques~\cite{Akihiko2019Recent}. Several approaches exist for the fabrication of vision-based tactile sensors. However, in most occurrences and although these sensing devices are equipped with camera(s), the visual perception is normally limited to the sensor's internal chamber. 

Vision-based tactile sensors generally employ at least one camera that captures the deformation of an elastomer while external pressure is exerted on its surface. From the images captured by the camera, the goal is usually to create a depth map that reflects the penetration of the object into the elastomer, to determine the force distribution on the sensor, or to infer a light reflection-to-pressure mapping. To meet such objectives, some have relied on frustrated total internal reflection (FTIR) by using an elastomer sitting on a light-conductive material~\cite{Begej1988Planar,Ohka2008Experimental}. In this setting, the elastomer has to be non-transparent to generate a perceptible reflection to the camera, therefore limiting the camera's view to the sensor's internal chamber. Other approaches involve tracking pins~\cite{Ward2018TacTip}, scattered particles~\cite{Sferrazza2019Design} or markers~\cite{Yamaguchi2017Optical} displacement, which can be used to measure normal force, shear and torque. However, the presence of such particles in the elastomer generally hinders efforts for external perception. Another popular principle in the literature is photometric stereo~\cite{Robert1980Photometric}, which consists of generating a depth map by illuminating the elastomer coated with reflective pigments from different directions. Following this method, Gelsight sensors~\cite{Johnson2009Retrographic,Donlon2018GelSlim}, Digit~\cite{Lambeta_2020} and others~\cite{Ozdemir2022HySenSe,Yuan2017GelSight} use distinct red, green and blue lighting from different directions to illuminate an elastomer that allows the generation of a depth map from a single image. Using photometric stereo with markers have also been proposed~\cite{Yuan2017GelSight,Shaoxiong2021GelSight}, which allow geometric reconstruction along with the determination of force distribution on the contact medium. Classical stereophotometric approaches usually involve the application of a non-transparent specular or matte coating on the contact medium, which limits the camera's perception to the inside of the sensor. Other types of vision-based tactile sensors have been proposed, such as stereoscopic-based sensors that get 3D images of a membrane deformation without constrained illumination~\cite{Jingyi2023GelStereo,Shaowei2022InHand}, sensors using a mix of photometric stereo and structured light~\cite{Huanbo2021soft}, and sensors based on illumination contrast and difference images~\cite{Changyi2022DTact}. However, all of the aforementioned sensors also suffer from the same limitation of using camera(s) that are able to capture only the inside portion of the sensor.\vspace{-0.1in}

\subsection{Vision-Based Sensors with Proximity/3D Geometry Sensing}
Nevertheless, a sensor that could combine both tactile and external vision would be a valuable asset for a wide variety of tasks, such as path and grasp planning, pose estimation, object reconstruction and many more. For this reason, researchers have sought ways to integrate both of these sensing modalities in a compact device.\comments{Gao et al.~\cite{Yuan2023InHand} used two external cameras along with two Gelsight sensors mounted on the same gripper to estimate 6D poses of grasped objects. However, the method required installing the sensors at four different locations on the gripper, which could increase the complexity of implementing this solution in terms of required space, wiring management and communication with devices each located at distinct positions. Therefore, an interesting alternative would consist of using a vision-based sensor capable of sensing both tactile and visual stimuli from within the same device.} One well-known example of such sensor is Fingervision~\cite{Yamaguchi2017Implementing}, which combines tactile sensing through marker tracking with 2D vision. By analyzing multiple 2D images with optical flow, the authors were able to estimate the proximity of objects. Fingervision have been successfully used for cutting vegetables~\cite{Yamaguchi2016Combining}, however, the markers printed on the external layer create noise in the images and their use of a single camera limited perception to 2D. Conversely, a rare occurrence where a single sensor combines both tactile sensing with 3D vision is the work of Shimonomura et al.~\cite{Shimonomura2016Robotic}, which combines FTIR sensing with stereoscopic vision. However, the sensor used FTIR technology with a rigid acrylic surface as the contact medium, which means it only provides information on the position and area of contact without capturing grasping forces or enabling contact imprint reconstruction. On the other hand, Hogan et al.~\cite{Hogan2021Seeing} showed that it was possible to see through a semi-reflective coating, used for photometric stereo by varying the sensor's internal lighting conditions\comments{illuminance?}. The authors have demonstrated the feasibility of estimating object proximity using an approach that required artificial vision techniques and moving the robot on a predetermined linear path to acquire images~\cite{Hogan2022Finger}. However, this method used a single camera, which prevented the sensor from perceiving the 3D geometry of objects. Recently, Luu et al.~\cite{Luu2022Soft} introduced ProTac, a sensor with a contact medium made of polymer dispersed liquid crystal (PLDC) film with markers, which allowed the active control of the membrane's transparency.\comments{Being able to actively alter the interface's transparency allowed the authors to acquire clear external perception and estimate proximity with a deep neural network.} However, the markers generated noise in the 2D images and the two cameras integrated by the authors were facing each others in the current implementation, which would not be well-suited for acquiring 3D features. 

In order to combine tactile and visual data, we rather show the development of a novel visuotactile sensor that combines 3D vision with tactile sensing. The sensor relies on stereoscopic vision for 3D imagery and photometric stereo, that benefits from using an adjustable-transparency-through-lighting elastomer, for tactile sensing.\vspace{-0.1in}

\section{Design and Sensing Principles} \label{sec:DesignFabrication}\vspace{-0.1in}
\subsection{Fabrication}
\begin{figure}[t]
      \centering
      \includegraphics[width=1\linewidth]{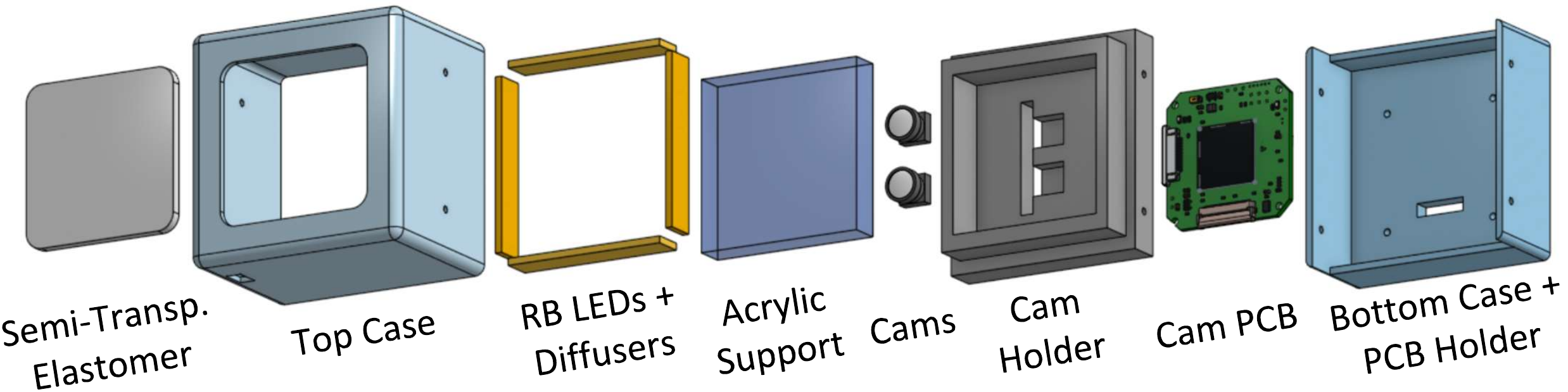}\vspace{-0.1in}
      \caption{Exploded view of the proposed sensor.}
      \label{fig:explodedView}\vspace{-0.25in}
    \end{figure}

\vspace{-0.05in}
    The design objective is to create a visuotactile sensor that integrates stereoscopic vision to capture a 3D representation of the nearby environment before contact and uses photometric stereo to acquire a 3D depth map (tactile imprint) of the contact. This capability could complement existing tactile-based manipulation approaches in cluttered environments (e.g.:~\cite{Thomasson2022Going}), where the ability to accurately perceive the surrounding environment and the object being manipulated is often crucial. Fig.~\ref{fig:explodedView} provides an exploded view of the elements used in the proposed design. The design involves several key components, including two 2D cameras, the contact elastomer and the semi-transparent layer. \comments{Each component plays a critical role in enabling the sensor to perform 3D reconstruction of objects and their surroundings. }In the following subsections, we describe how these elements were fabricated and assembled, along with specific design considerations.\vspace{-0.025in}
    
    \subsubsection{Elastomer and Coatings}
\begin{figure}[b]
          \centering\vspace{-0.2in}
          \includegraphics[width=\linewidth]{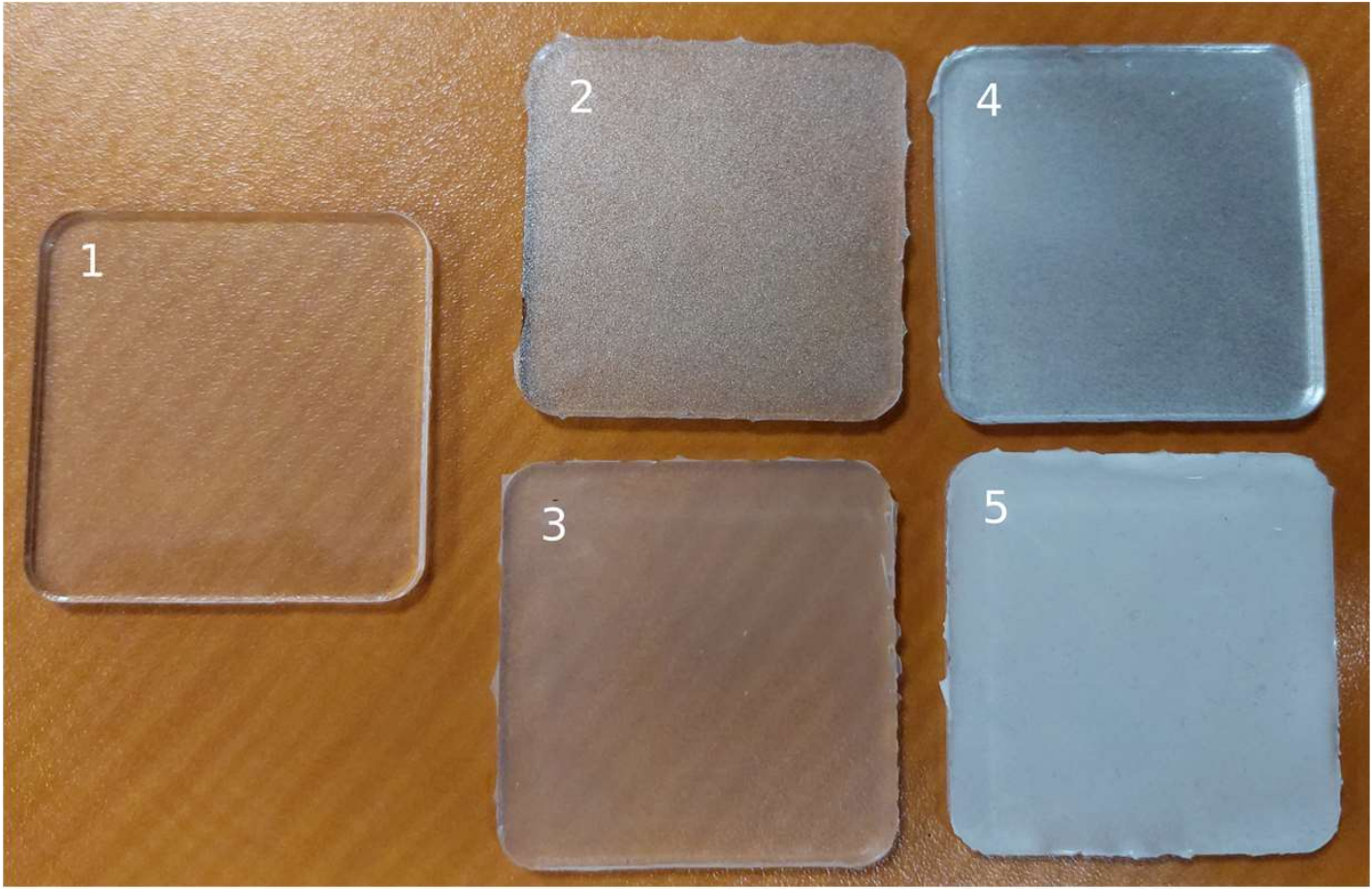}\vspace{-0.1in}
          \caption{The elastomers used in the comparative study: 1) Completely transparent, 2) Semi-transparent reflective, 3) Semi-transparent matte, 4) Opaque reflective, and 5) Opaque matte. It is important to note that membranes 1-3 are the focus of the study, while membranes 4-5 are not see-through and only used for comparison purposes in section \ref{sec:tactileSensing}.}\vspace{-0.15in}
          \label{fig:membranes}
        \end{figure}

        \comments{Elastomers were first investigated for their ability to deform upon encountering external pressure, resulting in modifications to the light trajectory emitted within the sensor and captured by the cameras.} As per prior studies \cite{Hogan2021Seeing, Donlon2018GelSlim}, P-595 silicone elastomer from Silicones Inc. was employed as the base material for all considered membranes, as depicted in Fig.~\ref{fig:membranes}. The elastomers were produced using a mold made of three 1/8 inch acrylic sheets,  with one sheet cut to the desired membrane shape and the remaining two employed to compress the gel and flatten the surfaces.

        The transparent and semi-transparent reflective membranes studied in this paper, which correspond to Fig.~\ref{fig:membranes}-\#1 and Fig.~\ref{fig:membranes}-\#2, were designed in accordance with the specifications outlined in \cite{Hogan2021Seeing}. To achieve the desired transparency, multiple layers of mirror-type spray paint (Rust-Oleum 267727) were applied onto the elastomer. A silicone protective layer was then sprayed over the paint layer to safeguard it. To create the protective layer, the same gel used for the membrane was mixed with a silicone thinner (Smooth-On NOVOCS Gloss) in a 2:1 ratio, resulting in a highly fluid gel that could be easily spread over the paint layer.

        The semi-transparent matte membrane (Fig.~\ref{fig:membranes}-\#3) was generated by incorporating 0.5\% white silicone dye (Smooth-On Silc Pig White) by weight into the protective layer. Subsequently, the opaque membranes (Fig.~\ref{fig:membranes}-\#5 and Fig.~\ref{fig:membranes}-\#5) were produced by adding 3\% white dye by weight to the protective layer. To create the reflective opaque membrane, a mirror paint layer was applied, similar to that of the semi-transparent reflective membrane, prior to coating with the opaque protective layer. The opacity of each membrane depicted in Fig. 4 was approximated using a stable light source and a lux meter (AP-881D from AOPUTTRIVER) positioned at a fixed distance. The sensing portion of the lux meter was covered by the membrane placed on an adapter, which only allowed light to pass through the membrane to reach the lux meter. The lux reading obtained in this manner and estimated opacity are shown in Table~\ref{tab:Opacity}. \vspace{-0.2in}

\begin{table}[h!]
    \centering
        \caption{Approximated Opacity for Each Membrane}\vspace{-0.1in}
    \begin{tabular}{|c|c|c|}\hline
        \textbf{Membrane Type} & \textbf{Total Light (Lux)} & \textbf{Approx. Opacity (\%)} \\\hline
         No membrane & 466 & n/a \\\hline 
         Transparent & 442 & 5.15 \\\hline 
         Semi-Trans. Reflective & 352 & 24.46 \\\hline 
         Semi-Trans. Matte & 363 & 22.10 \\\hline 
         Opaque Reflective & 93 & 80.04 \\\hline 
         Opaque Matte & 141 & 69.74 \\\hline 
    \end{tabular}\vspace{-0.1in}
    \label{tab:Opacity}
\end{table}

    \subsubsection{Lighting and acrylic support}
        For the internal illumination of the sensor, NeoPixel LED strips from Adafruit were positioned on the walls of the sensor enclosure at the same level as the acrylic membrane support. This configuration provides parallel illumination of the membrane surface, effectively reducing light leakage from the sensor. In particular, when using transparent or semi-transparent membranes, we aim to minimize light escaping from the sensor to prevent premature object illumination before contact. Further discussion on this topic can be found in section \ref{subsec:Discussion}. These lights are operated using an Arduino Micro communicating with a ROS Melodic package that we developed for StereoTac.

        To minimize internal reflections of the lights on the acrylic support block, a gray filter (VViViD Smoke Black Gloss Vinyl) was applied to the sides, as demonstrated in \cite{Shaoxiong2021GelSight}. Additionally, a 1624K57 LED diffuser from McMaster-Carr is employed to promote light diffusion instead of specularity.

    \subsubsection{Cameras}
        The prototype uses Odseven 160° variable Focus cameras, which offer a large field of view to capture the entire membrane. The cameras are particularly well-suited to this research due to their depth of field, which enables the focus to be on the membrane deformations while also capturing objects up to approximately 60 cm in front of the sensor. These cameras are operated by an Arducam USB3.0 Camera Shield capable of simultaneously acquiring images from two cameras. However, the cameras' circuitry required a modification for stereoscopic usage. Specifically, to synchronize the camera images, the clock component was removed from one camera's electrical circuit and that camera was connected to the clock of the second camera. This ensured precise synchronization between the cameras, allowing simultaneous image acquisition by the acquisition card. The cameras are positioned 14 mm apart, providing better depth resolution for close-up shots near the sensor, but decreasing resolution for more distant objects, which is not a focus of this study.\vspace{-0.1in}

\subsection{Stereoscopic Vision}\label{sec:Stereo}\vspace{-0.05in}
 \comments{   Stereoscopic vision is a well-established technique in robotics to obtain a 3D representation of an environment. In the context of visuotactile sensing, stereoscopic vision plays a crucial role in enabling the accurate 3D reconstruction objects located close to the sensor. In StereoTac design, the cameras are positioned such that a baseline separation is created between them, which allows for the computation of depth information. The images from the two cameras are captured simultaneously, which ensures that the images are synchronized and aligned correctly for 3D reconstruction.} 

    \comments{To provide the sensor with stereoscopic vision capabilities, two cameras are positioned 14 mm apart.} 
    Although the 14 mm spacing between the cameras limits the depth perception resolution at longer distances, it allows the sensor to be more compact and to perceive objects in a closer range, which is helpful for fine manipulation and more suited to confined and/or cluttered spaces (e.g.:~\cite{Thomasson2022Going}). To calibrate both the intrinsic and extrinsic parameters of the cameras, the stereo camera calibration utility available in the camera\_calibration package~\cite{cameracalibration2010} on ROS was used, given its ease of use and its visual feedback during the calibration process. The calibration was performed using an 8x6 checkerboard with 17mm tiles. The information obtained from calibration enables image rectification. Indeed, due to the cameras having wide-angle lenses, the initial images are distorted and require rectification prior to utilization. Using the Q Matrix obtained from calibration, the Stereo Block Matching utility provided by OpenCV is utilized to generate the disparity map and 3D projection of the scene. Finally, the obtained points are filtered by statistically removing outliers using the Open3D~\cite{Zhou2018} library before publishing the final point cloud.\vspace{-0.1in}

\subsection{Tactile Sensing}\label{sec:tactileSensing}\vspace{-0.05in}
    In tactile mode, the visuotactile sensor utilizes photometric stereo to capture the deformation of the membrane in response to the pressure and force applied by an object. Similar to other tactile sensors such as GelSight \cite{Yuan2017GelSight}, membrane deformation is captured using arrays of LEDs to illuminate the membrane from multiple angles, and a camera is then used to reconstruct the 3D shape of the object.

    \comments{StereoTac features a semi-transparent membrane that allows both the deformation of the membrane and the visual information of the object to be captured.} To ensure that only the deformation of the membrane is captured while StereoTac is operated in tactile mode, the exposure of the cameras is reduced and the LED arrays inside the sensor are illuminated to make the luminosity high enough to capture reflections. Moreover, to eliminate any potential contamination from external lighting in tactile mode, a HSV filter is applied to isolate only the red and blue tones corresponding to the LEDs' illumination. The LED arrays are placed on the four sides of the square-shaped lens, providing a range of angles to illuminate the membrane and enabling the detection of fine surface details. When the goal is only to obtain the shape of the contacts on the membrane, all the LED arrays can be illuminated to observe the reflection on the membrane's surface with the camera. However, when the goal is to perform 3D reconstruction of the contact on the membrane, directed light can be used to capture the reflection gradient and create a 3D representation of the tactile imprint.

    The following sub-sections present the 3D contact reconstruction method with StereoTac. The approach involves capturing contact gradients on a membrane utilizing photometric stereo technique. Additionally, we will describe the sensor calibration procedure, specifically regarding its ability to read gradients on a semi-transparent membrane. 

        \begin{figure}[t!]
            \centering
            \includegraphics[width=\linewidth]{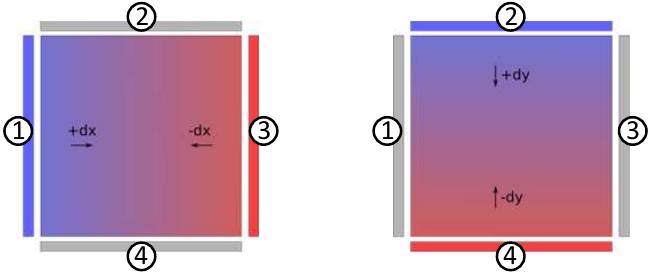}\vspace{-0.15in}
            \caption{The 2-step gradient capture method. 1) Gradient dy is obtained by illuminating the membrane with LED array 2 and 4 with blue and red respectively. 2) Gradient dx is obtained by illuminating the membrane with LED array 1 and 3 with blue and red respectively.}\vspace{-0.2in}
            \label{fig:TouchDxDyExplained}
        \end{figure}

    \subsubsection{\textbf{Capturing contact gradients from membrane illumination}}\label{sec:3dTouch}
\begin{figure}[b!]
            \centering\vspace{-0.15in}
            \includegraphics[width=\linewidth]{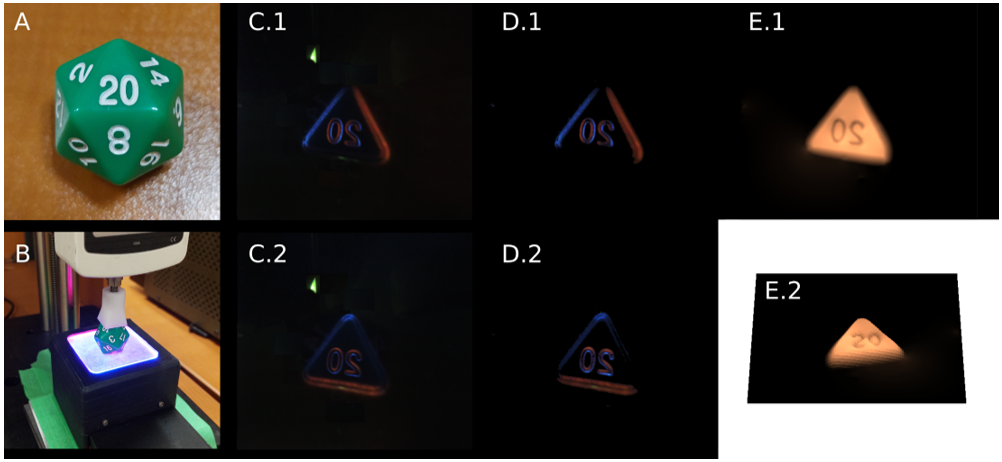}\vspace{-0.15in}
            \caption{Process for the 3D reconstruction of contacts. A) The example-object (D20). B) D20 contacting the reflective semi-transparent membrane. C.1-2) Images obtained during the two-step illumination dx-dy: note that the white/green spot is a ceiling light in the room where the image was taken (339 Lux). D.1-2) Images obtained after filtering C.1-2 using a color filter. E1-2) 3D reconstruction obtained with gradients calculated using images D.1-2.}
            \label{fig:TouchStep}
        \end{figure}

        \comments{Photometric stereo is a technique that involves capturing images of illuminated surfaces using different angles of illumination [CITE]. This approach allows determining the gradient of illuminated surfaces, and eventually constructing a three-dimensional object. The ability to take an image where membrane illumination is done through different color channels (RGB) at different angles of incidence is demonstrated by Gelsight \cite{Yuan2017GelSight}.}
        By utilizing the reflectance map of the elastomer, it is possible to determine the gradient at each pixel based on the intensity of the measured colors. Regarding the StereoTac sensor, the membrane is illuminated \comments{perpendicularly to it, }by placing the sensor lights at the elastomer's perimeter in a square configuration, as displayed in Fig.~\ref{fig:explodedView}. This setup provides four feasible illumination angles. However, a three-angle approach using RGB colors on three distinct axes, as is frequently done with photometric stereo-based sensors, would not be sufficient to observe the complete contact gradients. Specifically, the utilization of three colors simultaneously on three different, orthogonal angles would only provide a fraction of the information in the fourth direction where no directed light is employed. To address this limitation, we utilized the placement of the illumination axes to illuminate the elastomer in two steps, as shown in Fig.~\ref{fig:TouchDxDyExplained}. Given the parallel arrangement of the LED rows, the x-gradient (dx) information is obtained by illuminating rows 1 and 3 with distinct colors. Similarly, the y-gradient (dy) information is obtained from rows 2 and 4. Since illumination is done from only two directions at the same time, we employed blue-red pairs to facilitate color segmentation during image processing. The dx and dy gradients were captured sequentially by alternating the illuminated LED pairs, with dx and dy LEDs alternating at a rate of 4 Hz. Image capture from the cameras was synchronized to obtain two distinct images of membrane illumination. An example of the resulting images for the semi-transparent reflective membrane (Fig.~\ref{fig:membranes}-\#2) can be observed in Fig.~\ref{fig:TouchStep}.

        \comments{Following acquisition of the dx and dy gradient images of contacts on the membrane, we apply a mask to extract relevant information. As we are testing the use of transparent and semi-transparent membranes,} After acquiring the dx and dy gradient images, an HSV filter is applied to retain only the pixels corresponding to the red and blue tones of the LEDs used. This step eliminates potential contamination captured by the cameras originating from external lighting.\comments{To achieve this, we create a mask using the HSV values of the images to exclude any information that does not contain the colors of our LEDs.} As different membranes are investigated during this study, and their exact reflectance maps are unknown, a simple, 3-hidden layers neural network is employed, as proposed by Wang et al.~\cite{Shaoxiong2021GelSight}. Calibration is performed for each membrane by capturing 30 images, with a calibration ball of known diameter pressed at various positions on the membrane. After using the HSV mask to eliminate non-contact information from the image, and by knowing the diameter of the ball and the exact center position of the ball in the image, the dx and dy values of each pixel in contact with the ball can be easily determined using a simple distance relationship with its center. For example: $d_x = \sin^{-1}((p_x - c_x)/r )$, where $p$ is the position of the pixel, $c$ is the position of the center of the ball in pixels and $r$ is the radius of the ball in pixels.

        All non-zero pixels obtained from this acquisition are utilized to train the MLP neural network. The input to the network consists of the R-B values and the x-y position of each pixel. Including x-y values aims to mitigate the light attenuation problem, where pixels further away from the light source have lower brightness. This issue arises because, for the same contact gradient, the red brightness of a pixel closer to the red light source is higher than that of a pixel on the opposite side of the membrane.

        \begin{figure}[t]
            \centering
            \includegraphics[width=0.95\linewidth]{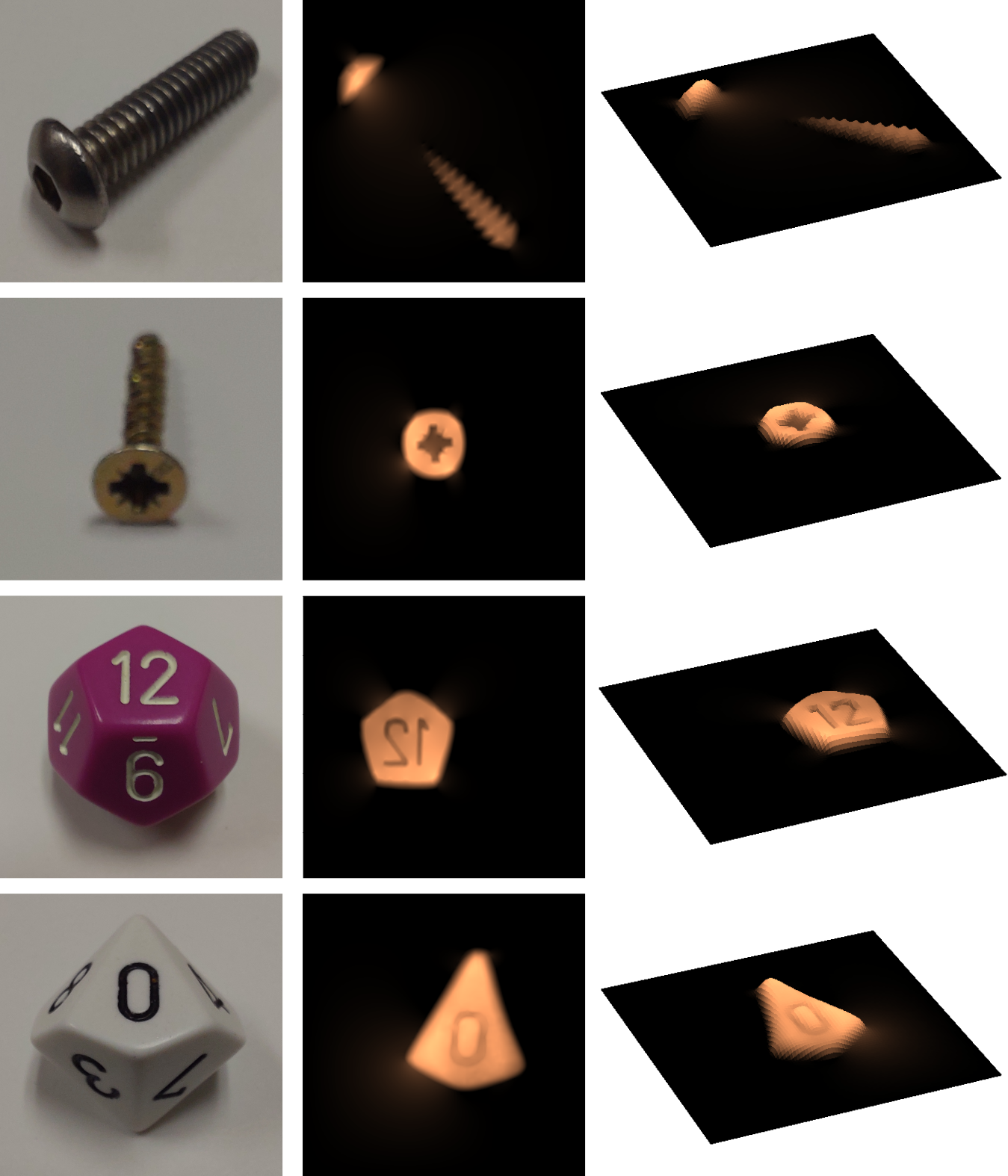}\vspace{-0.1in}
            \caption{Examples of 3D contact reconstruction made with the semi-transparent reflective membrane. Left, the real objects (M5 screw, wood screw, D12 and D0). Center, top view of the 3D imprints. Right, isometric view of the imprints.}
            \label{fig:touchExamples}
            \vspace{-0.25in}
        \end{figure}

    \subsubsection{\textbf{Reconstructing the 3D Tactile Imprint using gx-gy}}
        As demonstrated by Wang et al.~\cite{Shaoxiong2021GelSight}, it is possible to use a 2D Fast-Poisson solver to compute the depth ($z$) of each pixel given the gx and gy values of each pixel from an image capture. This method is particularly useful when dealing with noisy or incomplete gradient data, as it generally produces smoother results during 3D reconstruction. In our case, as the gradients are estimated by our neural network, we found that this reconstruction method provided a more accurate tactile imprint with finer details. Finally, by knowing the number of pixels per mm in the obtained images (in our case, 15 pixels/mm), we can determine the measured depth in millimeters. We used the open-source Python code from Doerner~\cite{doerner} to compute the Fast Poisson algorithm. Fig.~\ref{fig:touchExamples} provides an overview of the reconstruction process for several objects with different shapes.\vspace{-0.1in}

\section{Experiments}\vspace{-0.05in}
    \subsection{Evaluation of Visual Depth through different membranes}\vspace{-0.05in}
        We assess the depth perception capability by utilizing various evaluation metrics. In particular, the performance of the sensor in vision mode is evaluated using \textit{Z-accuracy}, \textit{RMS error} (spatial noise) and \textit{temporal noise}. These metrics are commonly used in the evaluation of depth cameras and correspond to some of the recommended metrics in the "Camera Depth Testing Methodology" from Intel \cite{intelTest}. 

        To perform empirical experiments on the membranes (\#1, \#2 and \#3 in Fig.~\ref{fig:membranes}), we positioned the sensor at different distances (10 cm, 15 cm, 20 cm, 25 cm, and 30 cm) away from a completely flat surface.\comments{ By performing this analysis on multiple images and distances, the experiment provides important insights into the spatial noise and precision of the depth measurements. This information is essential for evaluating the accuracy and reliability of the sensor and for identifying any sources of error or variability in its performance.} At each position, ten RGB-D images were captured with each of the membranes. Additionally, the same experiments were repeated using the Intel RealSense D405 camera, which is an off-the-shelf short-range stereoscopic camera, for comparison purposes.


    \subsubsection{\textbf{Z-accuracy}} 
\begin{figure}[h]
        \centering\vspace{-0.1in}
        \includegraphics[width=0.95\linewidth]{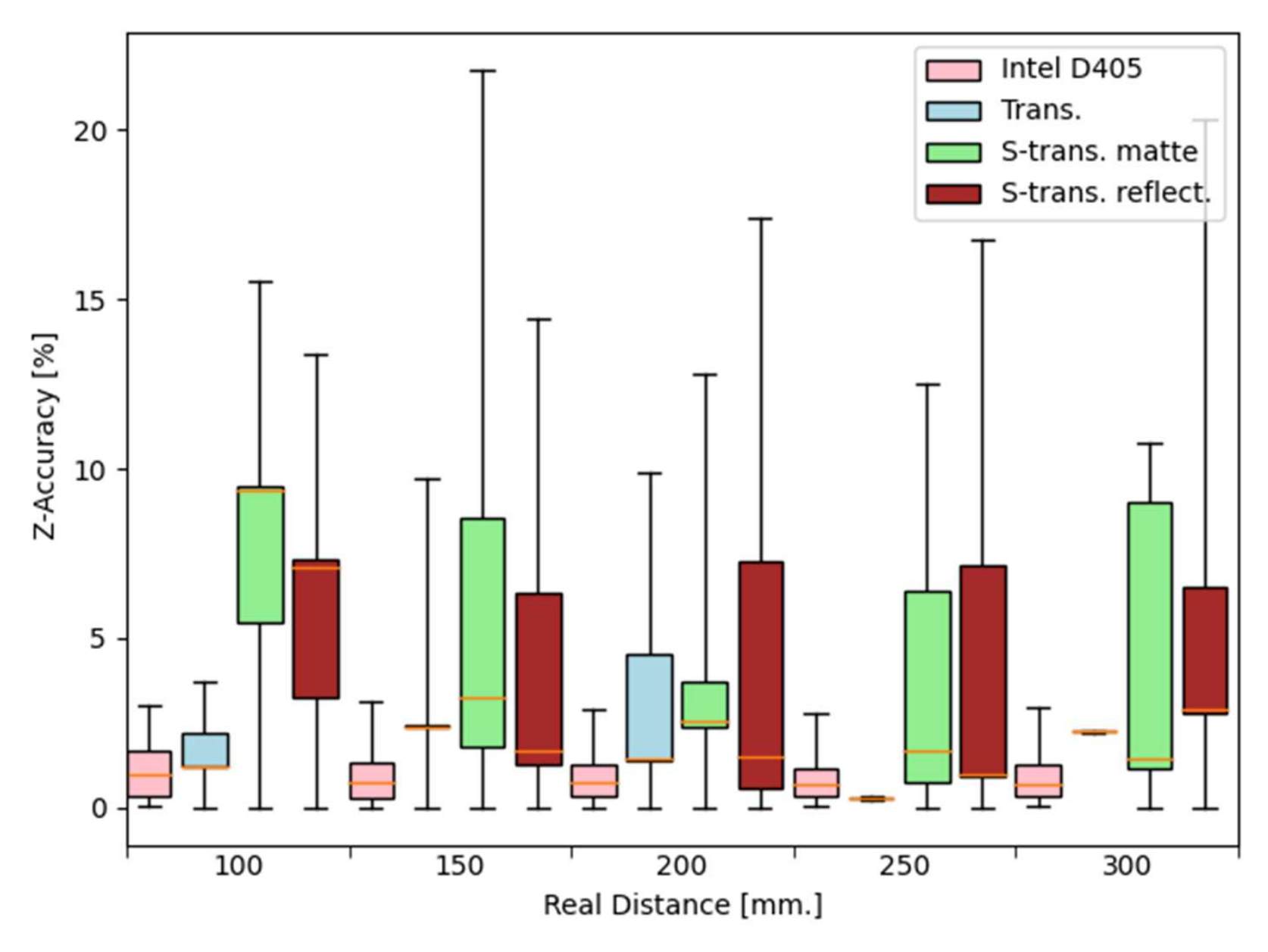}
        \vspace{-0.2in}
        \caption{Z-Accuracy measured on a flat surface using different membranes at different distances.}
        \label{fig:zaccuracy}
        \vspace{-0.05in}
    \end{figure}

Z-accuracy measures the depth accuracy by taking a pre-measured ground truth as a reference. The metric expresses the accuracy of depth data on a per-pixel basis relative to the ground truth for each captured depth image at a fixed distance. The depth is calculated relative to a best-fit plane in the point cloud to isolate camera positioning errors. The Z-accuracy is then obtained by calculating 
    \begin{equation}
    Z_{acc} = \frac{median(D(x,y)-GT)}{GT}\times 100\%,    
    \end{equation}
    where $D(x,y)$ is the calculated depths at pixel positions $(x,y)$ and GT is the ground-truth depth value. The resulting Z-Accuracy is then a ratio of the error and the actual distance. The experiment results can be observed in Fig.~\ref{fig:zaccuracy}.
    
    \comments{The utilization of a semi-transparent membrane has been observed to influence the precision of depth measurement, while the impact of matte or reflective finishes on this aspect creates more variations in depth perception as well as a less accurate median value in the distribution, when compared to the Intel camera and in the case of using a transparent membrane.}

    The precision of depth measurement was found to be affected by the use of a semi-transparent membrane. As expected from the presence of distortions created by using a semi-transparent interface, the presence of matte or reflective finishes created more variations in depth perception and resulted in a generally less accurate median values ($\sim 0.5-9\%$) in the distribution, when compared to the Intel camera or StereoTac equipped with a transparent membrane.
    
    \subsubsection{\textbf{RMS error}} The Root Mean Square (RMS) error and spatial noise of StereoTac's depth measurements are evaluated. While the variation in depth values of each pixel within a ROI does not directly measure accuracy, it is an important metric for assessing the quality of depth measurements, as it provides information on the consistency and repeatability of the measurements within a specific area. The results obtained during this experiment can be observed in table \ref{tab:RMSE}.\vspace{-0.1in}
    

        \begin{table}[hb]
        \centering\vspace{-0.05in}
        \caption{RMSE Values obtained on a Flat Surface by Different Membranes at Different Distances, (\%) [$\mu : \sigma$]}\vspace{-0.1in}
        \begin{tabular}{l|llll}
        \textbf{Dist.} & Intel D405  & Transparent & \begin{tabular}[c]{@{}l@{}}Semi-trans.\\ matte\end{tabular} &  \begin{tabular}[c]{@{}l@{}}Semi-trans.\\ reflective\end{tabular} \\ \hline
        10 cm.             & 0.43 : 0.01 & 4.1 : 1.66  & 2.18 : 0.1                                                      & 8.06 : 1.43                                                          \\
        15 cm.             & 0.6 : 0.02  & 2.3 : 0.7   & 3.52 : 0.59                                                     & 4.70 : 0.79                                                           \\
        20 cm.             & 0.86 : 0.03 & 2.23 : 0.06 & 5.91 : 3.19                                                     & 6.65 : 0.48                                                           \\
        25 cm.             & 1.01 : 0.04 & 1.92 : 0.52 & 3.08 : 0.06                                                     & 4.76 : 0.43                                                           \\
        30 cm.             & 1.15 : 0.02 & 1.89 : 0.18 & 3.74 : 0.13                                                     & 6.15 $\pm$ 1.35                                                           
        \end{tabular}
        \label{tab:RMSE}
        \end{table}\vspace{-0.1in}

    Table \ref{tab:RMSE} reveals that the reflective semi-transparent membrane exhibits a higher RMSE error, consequently accounting for the higher Z-Accuracy error. The precision and reliability of the whole system are indeed dependent on the intrinsic variation of measurements on a single image, explaining the correlation between the RMSE and Z-Accuracy values.
    
    \subsubsection{\textbf{Temporal Noise}}  The changes in depth values for each pixel within the same ten images were monitored over time for each distance. To quantify the temporal noise, the standard deviation of depth values across the ten images of a flat surface was determined. The mean of the standard deviations of all pixels within the ROI was calculated and used as the measure of temporal noise. The results of this experiment can be observed in the table~\ref{tab:temporalNoise}.

        \begin{table}[ht]
        \centering\vspace{-0.1in}
        \caption{Temporal Errors Obtained on a Flat Surface by Different Membranes at Different Distances. [\%]}\vspace{-0.1in}
        \begin{tabular}{l|llll}
        \textbf{Dist.} & Intel D405 & Trans. & \begin{tabular}[c]{@{}l@{}}Semi-trans.\\ matte\end{tabular} & \begin{tabular}[c]{@{}l@{}}Semi-trans.\\ reflective\end{tabular} \\ \hline
        10 cm.            & 0.29       & 1.25        & 0.75                                                            & 3.72                                                                  \\
        15 cm.            & 0.46       & 0.81        & 1.42                                                            & 3.48                                                                  \\
        20 cm.            & 0.69       & 0.72        & 1.84                                                            & 3.12                                                                  \\
        25 cm.            & 0.81       & 1.17        & 1.02                                                            & 2.86                                                                  \\
        30 cm.            & 0.96       & 1.64        & 0.89                                                            & 4.57                                                                 
        \end{tabular}
        \label{tab:temporalNoise}\vspace{-0.1in}
        \end{table}

    The results show that the use of a matte semi-transparent membrane does not significantly affect the noise level over time. However, the use of the reflective semi-transparent membrane adds observable noise. For example, at 10cm, the temporal noise would be around $\pm$ 3.72 mm. This increase may be due to the clarity of the membrane, where the reflective semi-transparent membrane has tiny visible paint spots when viewed up close by the cameras.\vspace{-0.2in}

    \subsection{Tactile Imprint Experiments}\vspace{-0.05in}
        To determine the accuracy and stability of depth measurement when the sensor is in tactile mode, empirical experiments were conducted by pressing a known-sized flat object onto the sensor at a repeatable depth. To do this, a 13mm diameter disk mounted on a Mark-10 Manual Lever Operated Test Stand was used to ensure perpendicular pressure onto the membrane and capture a truly flat image. To ensure the penetration depth of the disk on the membrane, a Mitutoyo Absolute 543-693 vertical vernier, mounted on the test stand, was employed.
        
        To obtain a reliable estimation of the depth reconstruction error for each membrane, pressure disks were consecutively pressed onto the sensor 30 times at random locations to a depth of 1 mm. The methods discussed in section~\ref{sec:3dTouch} were then used to reconstruct the disk. Depth measurements for each trial were obtained by taking the average of the depth values of the flat surface of the disk in the image. Table~\ref{tab:touchResult} provides the mean and standard deviation values for each membrane.

        \begin{table}[h!]
            \centering \vspace{-0.1in}
            \caption{Mean and Standard Deviation of Depth Measurements\textemdash 1mm Indentation with a disk onto each membrane 30 times. [mm.]}\vspace{-0.1in}
            \begin{tabular}{lll}
            \textbf{Membrane type} & \textbf{Mean} & \textbf{Std}   \\ \hline
            Clear                  & 0.915         & 0.179          \\
            Semi-Reflective        & 0.837         & \textbf{0.085} \\
            Semi-Matte              & 0.6143        & 0.121          \\ \hline
            Opaque-Reflective      & 1.094         & \textbf{0.071} \\
            Opaque-Matte            & 0.853         & 0.091         
            \end{tabular}
            \label{tab:touchResult}
            \vspace{-0.15in}
        \end{table}

        The mean values are generally below 1mm for all membranes except for those obtained using the Opaque-Reflective membrane. This may be due to several factors. For example, since a neural network was employed to obtain depth measurements, the calibration process of the membranes plays a crucial role in determining the accuracy of the depth estimates. The calibration was performed using a sphere with a textured surface, and it is possible that the estimation of depth on the smooth surface of the disk altered the perspective of the measurements. However, it is worth noting that the two reflective membranes yielded the smallest standard deviations in measurement, indicating that depth measurement using them resulted in more consistent outcomes.\vspace{-0.1in}

\begin{figure}[htb!]
    \centering
    \includegraphics[width=\linewidth]{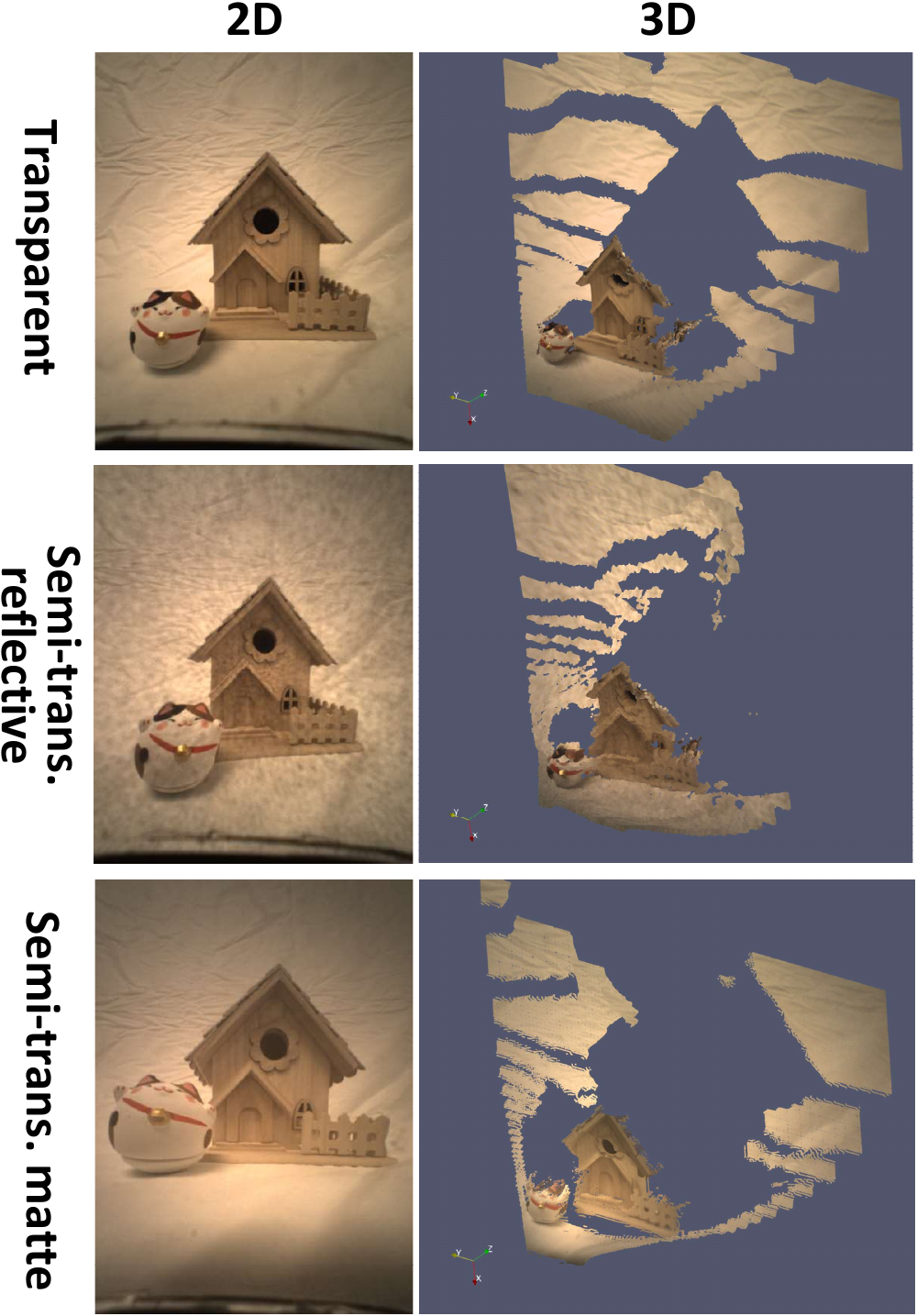}\vspace{-0.1in}
    \caption{Comparison of 2D and 3D perception through different membranes. For a more comprehensive understanding of the sensor's 3D vision capabilities, the reader is invited to refer to the \href{https://youtu.be/64nJgx2AMV4}{\textcolor{blue}{{\setul{1pt}{.4pt}\ul{accompanying video}}}}.}
    \label{fig:StereoMembraneComp}\vspace{-0.25in}
\end{figure}

\subsection{Discussion}\label{subsec:Discussion}\vspace{-0.05in}
    Fig.~\ref{fig:StereoMembraneComp} showcases a qualitative comparison of StereoTac's 3D vision capabilities using the three see-through membranes. Stereoscopic assessments have noted that semi-reflective membranes generally undermine depth measurement reliability. The errors detected are typically limited to a maximum of 2 centimeters over a distance of 30 cm. While these errors bear significance in the context of precise robotic grasping, real-time depth readings are still feasible to estimate the actual depth required to reach a target. Moreover, the utilization of stereoscopy for depth sensing is highly influenced by ambient brightness in the scene. Depending on the brightness variability of the task, integrating external LEDs to the sensor could adjust the ambient brightness for uniform readings. Additionally, the membrane's cleanliness can impact the depth reading's precision. For example, handling oily or dirty objects may leave residue on the membrane, which could potentially hinder reliable readings.

    Concerning the tactile properties of the membranes, it can be qualitatively observed from Fig.~\ref{fig:touchExamples} that semi-transparent membranes offer 3D reconstructions suitable for object recognition during grasping. The empirical results of experiments presented in Table \ref{tab:touchResult} indicate that the use of reflective semi-transparent membranes provides the best reconstruction reliability for our sensor type.

\begin{figure}[h!]\vspace{-0.1in}
            \centering
            \includegraphics[width=0.7\linewidth]{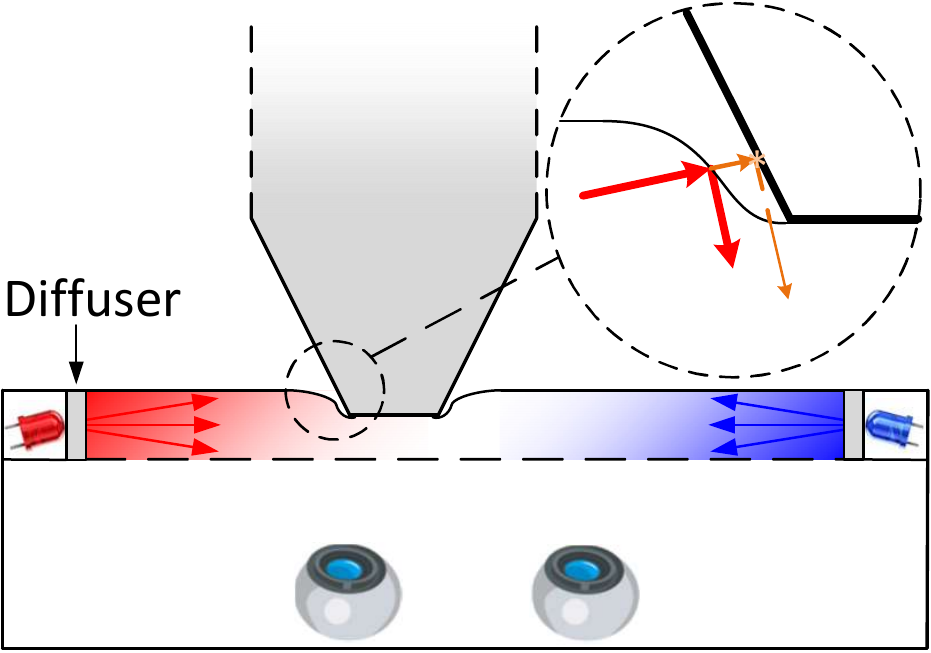}\vspace{-0.1in}
            \caption{The effect of semi-transparent / transparent coating on a soft elastomer. In the top-right corner: for an elastomer coated with opaque reflective paint, only the red light ray will be reflected to the camera, while only the orange light ray will be seen when a completely transparent membrane is used. When the coating is semi-transparent, both red and orange rays will be seen by the camera.}\vspace{-0.1in}
            \label{fig:AmbiguousReflection}
\end{figure}

    However, it is important to note that the use of transparent or semi-transparent membranes comes with noise effects that would not be present with opaque membranes. A fraction of the sensor's internal illumination escapes through the semi-transparency of the membrane, and depending on the object's color, incident angle or reflectivity, depth readings may be altered. The red or blue color rays emitted from the sensor can reflect off the object and return to the sensor, marginally modifying the illumination readings, as exemplified in Fig.~\ref{fig:AmbiguousReflection}.

\begin{figure}[t]
                \centering\vspace{-0.05in}
                \includegraphics[width=0.9\linewidth]{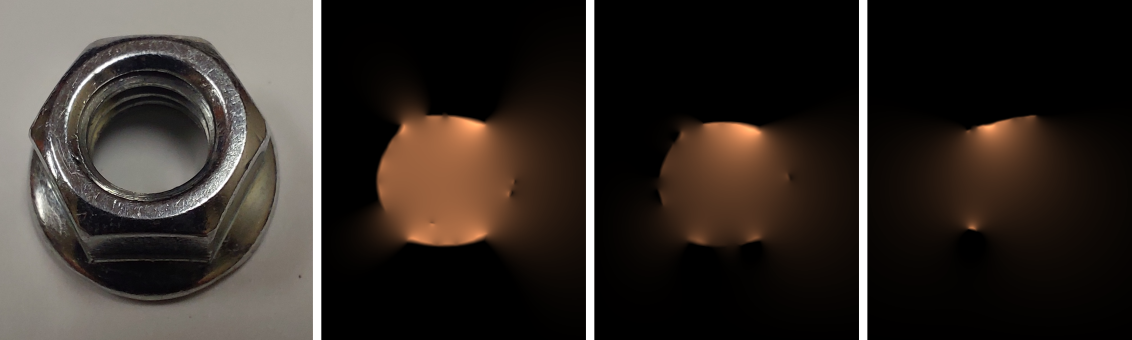}\vspace{-0.1in}
                \caption{Example of contact measuring error with a reflective object held at 1mm. from the membrane. 1) Reflective bolt. 2) Transparent membrane. 3) Semi-transparent opaque membrane. 4) Semi-transparent reflective membrane.}
                \label{fig:1mmOutExample}
                \vspace{-10pt}
\end{figure}

    This phenomenon was created on purpose and is observable in Fig.~\ref{fig:1mmOutExample}. In this scenario, we suspended a reflective bolt 1 mm above the membrane, which corresponds to the longest distance away from the sensor at which gradients could be perceived. The red and blue lights emitted from the sensor hit the reflective surfaces of the bolt and returned to the sensor, causing an inaccurate measurement. The transparent membrane is obviously more susceptible to this issue, but semi-transparent membranes are also slightly affected.

\section{Conclusion}
    This article presented the development of StereoTac, a novel tactile sensor with 3D vision capabilities. The feasibility of combining multiple modalities for robotic manipulation in a single sensor was demonstrated by incorporating a second camera inside a visuo-tactile sensor and using a semi-transparent contact membrane. Reliable 3D reconstruction of tactile imprints was shown to be achievable despite the semi-transparency of the membrane by capturing contact gradients in a sequential manner, as demonstrated by the results.
\comments{
    While this technology has promising potential for improving object grasping in cluttered environments, several areas of improvement need to be explored to fully realize its potential.}

    Nevertheless, several areas of improvement will be explored as future work to improve different aspects of the sensor. For example, tactile imprints were reconstructed using only one camera in this work. However, using the two available cameras of the sensor for this task could potentially increase the accuracy and confidence level of the reconstruction. Investigating the use of stereoscopy for contact reconstruction, as recently done in \cite{Jingyi2023GelStereo, Shaowei2022InHand}, would be interesting and promising.

    Furthermore, it was observed that the 3D view of the sensor environment was less reliable when using reflective semi-transparent membranes. Although no advanced preprocessing of the images from the two cameras was performed, it is possible that the use of image filtering as well as image reconstruction methods such as autoencoders could reduce the noise caused by the membrane and improve the depth estimates. Additionally, depth post-processing techniques like temporal filtering or edge-preserving filtering \cite{intelDepth} would be beneficial in reducing the point cloud distortions generated by semi-transparent membranes.\vspace{-0.15in}

\bibliographystyle{ieeetr}
\bibliography{main}

\begin{thebibliography}{10}

\bibitem{Yamaguchi2017Optical}
A.~Yamaguchi and C.~Atkeson, ``Optical skin for robots: Tactile sensing and
  whole-body vision,'' in {\em Carnegie Mellon University}, 07 2017.

\bibitem{Chi2018Recent}
C.~Chi, X.~Sun, N.~Xue, T.~Li, and C.~Liu, ``Recent progress in technologies
  for tactile sensors,'' {\em Sensors}, vol.~18, no.~4, 2018.

\bibitem{Shimonomura2019Tactile}
K.~Shimonomura, ``Tactile image sensors employing camera: A review,'' {\em
  Sensors}, no.~18, 2019.

\bibitem{Akihiko2019Recent}
A.~Yamaguchi and C.~G. Atkeson, ``Recent progress in tactile sensing and
  sensors for robotic manipulation: can we turn tactile sensing into vision?,''
  {\em Advanced Robotics}, vol.~33, no.~14, 2019.

\bibitem{Begej1988Planar}
S.~Begej, ``Planar and finger-shaped optical tactile sensors for robotic
  applications,'' {\em IEEE Journal on Robotics and Automation}, vol.~4, no.~5,
  pp.~472--484, 1988.

\bibitem{Ohka2008Experimental}
M.~Ohka, H.~Kobayashi, J.~Takata, and Y.~Mitsuya, ``An experimental optical
  three-axis tactile sensor featured with hemispherical surface,'' {\em Journ.
  of Adv. Mech. Des. Syst. and Man.}, vol.~2, pp.~860--873, 2008.

\bibitem{Ward2018TacTip}
B.~Ward-Cherrier, N.~Pestell, L.~Cramphorn, B.~Winstone, M.~E. Giannaccini,
  J.~Rossiter, and N.~F. Lepora, ``The tactip family: Soft optical tactile
  sensors with 3d-printed biomimetic morphologies,'' {\em Soft Robotics},
  vol.~5, no.~2, pp.~216--227, 2018.
\newblock PMID: 29297773.

\bibitem{Sferrazza2019Design}
C.~Sferrazza and R.~D’Andrea, ``Design, motivation and evaluation of a
  full-resolution optical tactile sensor,'' {\em Sensors}, vol.~19, no.~4,
  2019.

\bibitem{Robert1980Photometric}
R.~J. Woodham, ``{Photometric Method For Determining Surface Orientation From
  Multiple Images},'' {\em Opt. Eng.}, vol.~19, no.~1, 1980.

\bibitem{Johnson2009Retrographic}
M.~K. Johnson and E.~H. Adelson, ``Retrographic sensing for the measurement of
  surface texture and shape,'' in {\em 2009 IEEE Conference on Computer Vision
  and Pattern Recognition}, pp.~1070--1077, 2009.

\bibitem{Donlon2018GelSlim}
E.~Donlon, S.~Dong, M.~Liu, J.~Li, E.~Adelson, and A.~Rodriguez, ``Gelslim: A
  high-resolution, compact, robust, and calibrated tactile-sensing finger,'' in
  {\em 2018 IEEE/RSJ International Conference on Intelligent Robots and Systems
  (IROS)}, pp.~1927--1934, 2018.

\bibitem{Lambeta_2020}
M.~Lambeta, P.-W. Chou, S.~Tian, B.~Yang, B.~Maloon, V.~R. Most, D.~Stroud,
  R.~Santos, A.~Byagowi, G.~Kammerer, D.~Jayaraman, and R.~Calandra, ``{DIGIT}:
  A novel design for a low-cost compact high-resolution tactile sensor with
  application to in-hand manipulation,'' {\em {IEEE} Robotics and Automation
  Letters}, vol.~5, pp.~3838--3845, jul 2020.

\bibitem{Ozdemir2022HySenSe}
O.~C. Kara, N.~Ikoma, and F.~Alambeigi, ``Hysense: A hyper-sensitive and
  high-fidelity vision-based tactile sensor,'' 2022.

\bibitem{Yuan2017GelSight}
W.~Yuan, S.~Dong, and E.~H. Adelson, ``Gelsight: High-resolution robot tactile
  sensors for estimating geometry and force,'' {\em Sensors}, vol.~17, no.~12,
  2017.

\bibitem{Shaoxiong2021GelSight}
S.~Wang, Y.~She, B.~Romero, and E.~H. Adelson, ``Gelsight wedge: Measuring
  high-resolution 3d contact geometry with a compact robot finger,'' {\em
  CoRR}, vol.~abs/2106.08851, 2021.

\bibitem{Jingyi2023GelStereo}
J.~Hu, S.~Cui, S.~Wang, C.~Zhang, R.~Wang, L.~Chen, and Y.~Li, ``Gelstereo
  palm: a novel curved visuotactile sensor for 3d geometry sensing,'' {\em IEEE
  Transactions on Industrial Informatics}, pp.~1--10, 2023.

\bibitem{Shaowei2022InHand}
S.~Cui, R.~Wang, J.~Hu, J.~Wei, S.~Wang, and Z.~Lou, ``In-hand object
  localization using a novel high-resolution visuotactile sensor,'' {\em IEEE
  Trans. on Industrial Electronics}, vol.~69, no.~6, pp.~6015--6025, 2022.

\bibitem{Huanbo2021soft}
H.~Sun, K.~J. Kuchenbecker, and G.~Martius, ``A soft thumb-sized vision-based
  sensor with accurate all-round force perception,'' {\em CoRR},
  vol.~abs/2111.05934, 2021.

\bibitem{Changyi2022DTact}
C.~Lin, Z.~Lin, S.~Wang, and H.~Xu, ``Dtact: A vision-based tactile sensor that
  measures high-resolution 3d geometry directly from darkness,'' 2022.

\bibitem{Yamaguchi2017Implementing}
A.~Yamaguchi and C.~G. Atkeson, ``Implementing tactile behaviors using
  fingervision,'' in {\em 2017 IEEE-RAS 17th International Conference on
  Humanoid Robotics (Humanoids)}, pp.~241--248, 2017.

\bibitem{Yamaguchi2016Combining}
A.~Yamaguchi and C.~G. Atkeson, ``Combining finger vision and optical tactile
  sensing: Reducing and handling errors while cutting vegetables,'' in {\em
  2016 IEEE-RAS 16th International Conference on Humanoid Robots (Humanoids)},
  pp.~1045--1051, 2016.

\bibitem{Shimonomura2016Robotic}
K.~Shimonomura, H.~Nakashima, and K.~Nozu, ``Robotic grasp control with
  high-resolution combined tactile and proximity sensing,'' in {\em 2016 IEEE
  Int. Conf. on Rob. and Aut. (ICRA)}, pp.~138--143, 2016.

\bibitem{Hogan2021Seeing}
F.~R. Hogan, M.~Jenkin, S.~Rezaei-Shoshtari, Y.~Girdhar, D.~Meger, and
  G.~Dudek, ``Seeing through your skin: Recognizing objects with a novel
  visuotactile sensor,'' in {\em Proc. of the IEEE/CVF Winter Conf. on
  Applications of Computer Vision (WACV)}, pp.~1218--1227, January 2021.

\bibitem{Hogan2022Finger}
F.~R.~Hogan, J.-F. Tremblay, B.~H.~Baghi, M.~Jenkin, K.~Siddiqi, and G.~Dudek,
  ``Finger-sts: Combined proximity and tactile sensing for robotic
  manipulation,'' {\em IEEE R-A Letters}, pp.~1--8, 2022.

\bibitem{Luu2022Soft}
Q.~K. Luu, D.~Q. Nguyen, N.~H. Nguyen, and V.~A. Ho, ``Soft robotic link with
  controllable transparency for vision-based tactile and proximity sensing,''
  2022.

\bibitem{Thomasson2022Going}
R.~{Thomasson}, E.~{Roberge}, M.~{Cutkosky}, and J.~{Roberge}, ``Going in
  blind: Object motion classification using distributed tactile sensing for
  safe reaching in clutter,'' in {\em 2022 IEEE International Conference on
  Intelligent Robots and Systems (IROS)}, 2022.

\bibitem{cameracalibration2010}
``camera\_calibration - package summary.''
  \url{http://wiki.ros.org/camera_calibration}.
\newblock Accessed: 2023-03-06.

\bibitem{Zhou2018}
Q.-Y. Zhou, J.~Park, and V.~Koltun, ``{Open3D}: {A} modern library for {3D}
  data processing,'' {\em arXiv:1801.09847}, 2018.

\bibitem{doerner}
J.~Doerner, ``Fast poisson reconstruction in python.''
  https://gist.github.com/jackdoerner/b9b5e62a4c3893c76e4c.

\bibitem{intelTest}
Intel, ``Camera depth testing methodology.'' https://dev.intelrealsense.
  com/docs/camera-depth-testing-methodology, Jan 2021.

\bibitem{intelDepth}
Intel, ``Depth post-processing for intel realsense depth cameras.''
  https://dev.intelrealsense.com/docs/depth-post-processing, Jan 2021.

\end{thebibliography}

\end{document}